\documentclass{article}
\usepackage[inline]{enumitem}
\usepackage{graphicx}
\usepackage{amsmath}
\usepackage{tabularx}
\usepackage{makecell}
\usepackage[dvipsnames]{xcolor}
\usepackage[font=footnotesize,labelfont=bf]{caption}
\usepackage[draft]{hyperref}
\usepackage{natbib}
\usepackage{placeins}
\usepackage[T1]{fontenc}
\usepackage{CJKutf8}
\usepackage{wrapfig,lipsum,booktabs}
\usepackage[english]{babel}

\usepackage[accepted]{./sty/icml2017_arxiv}
\usepackage{multirow}
\usepackage{dblfloatfix}
\usepackage{slashbox}
\usepackage{makecell}
\usepackage{xcolor,colortbl}
\usepackage{amssymb}
\usepackage[cjk]{kotex}
\usepackage[normalem]{ulem}
\usepackage{multicol}
\usepackage{caption}
\usepackage{subcaption}

\icmltitlerunning{Paraphrase Thought: Sentence Embedding Module Imitating Human Language Recognition}

\begin{document}
	\begin{CJK}{UTF8}{mj}
		
		\twocolumn[
		\icmltitle{Paraphrase Thought: Sentence Embedding Module Imitating\\ Human Language Recognition}
		
		\begin{icmlauthorlist}
			\icmlauthor{Myeongjun Jang}{Ko}
			\icmlauthor{Pilsung Kang}{Ko}
		\end{icmlauthorlist}
		
		\icmlaffiliation{Ko}{School of Industrial Management Engineering, Korea University, Seoul, South Korea}
		\icmlcorrespondingauthor{Myeongjun Jang}{xkxpa@korea.ac.kr}
		\icmlcorrespondingauthor{Pilsung Kang}{pilsung\_kang@korea.ac.kr}
		
		\vskip 0.3in
		]
		
		\printAffiliationsAndNotice{}  

\begin{abstract}
Sentence embedding is an important research topic in natural language processing. It is essential to generate a good embedding vector that fully reflects the semantic meaning of a sentence in order to achieve an enhanced performance for various natural language processing tasks, such as machine translation and document classification. Thus far, various sentence embedding models have been proposed, and their feasibility has been demonstrated through good performances on tasks following embedding, such as sentiment analysis and sentence classification. However, because the performances of sentence classification and sentiment analysis can be enhanced by using a simple sentence representation method, it is not sufficient to claim that these models fully reflect the meanings of sentences based on good performances for such tasks. In this paper, inspired by human language recognition, we propose the following concept of semantic coherence, which should be satisfied for a good sentence embedding method: similar sentences should be located close to each other in the embedding space. Then, we propose the Paraphrase-Thought (P-thought) model to pursue semantic coherence as much as possible. Experimental results on two paraphrase identification datasets (MS COCO and STS benchmark) show that the P-thought models outperform the benchmarked sentence embedding methods.

\end{abstract}

\section{Introduction}
Sentence embedding, which transforms sentences into low-dimensional vector values reflecting their meanings, is a highly important task in natural language processing (NLP). By mapping unstructured text data into a certain form of structured representation, the embedding vector can enhance the performances of various NLP tasks, such as machine translation \cite{artetxe2017unsupervised,lee2016fully,zhao2016deep}, document classification \cite{conneau2017very,zhou2016text}, and sentence matching \cite{wan2016deep}. As sentence embedding plays an import role in NLP, various methods \cite{SkipThought,Sent2vec,FastSent,SIF,InferSent,Doc2vecC} have been proposed since the advent of the Doc2vec method \cite{Doc2vec}. Typically, these methods exhibit better performances than benchmarked embedding methods for common NLP tasks, such as document classification or sentiment analysis. However, this is not a direct evaluation of how well semantic meanings are preserved by the proposed embedding method.
	
Indirect methods for evaluating sentence embedding are not sufficient to evaluate the main property of sentence embedding techniques, i.e., how well semantic relationships between sentences are preserved. \citet{iyyer2015deep} showed that it is possible to achieve a fairly good performance in document classification using a simple document representation vector, i.e., an average of word vectors in the document. Even for classic document representation methods, in which word sequences or semantic relationships between words are not considered, e.g., bag of words (BoW) or term frequency-inverse document frequency (TF-IDF), highly accurate classification results can be achieved using a Na\"{i}ve Bayesian classifier \cite{soumya2278text}. This means that a good performance on a classification task can be achieved without the use of embedding vectors. In other words, a good classification performance for common NLP tasks using a certain type of sentence embedding method does not guarantee that the embedding method can successfully preserve the semantic relationship between sentences.

In this paper, in order to overcome the limitations of indirect sentence embedding evaluation strategies, we propose the following concept of semantic coherence, which should be satisfied by a good sentence embedding method: sentences having similar meanings should be placed close to each other in the embedding space. Then, we propose a new sentence embedding model named Paraphrase-Thought (P-thought), which can maximally pursue semantic coherence during training. The P-thought model is designed as a dual generation model, which receives a single sentence as input and generates both the input sentence and its paraphrase sentence simultaneously. The proposed P-thought model is evaluated through a task of measuring the semantic coherence and the STS Benchmark task. Experimental results show that the proposed P-thought model yields a better performance than benchmarked models in both tasks.
	
The remainder of this paper is organized as follows. In Section 2, we briefly review previous research on sentence embedding. In Section 3, we propose the concept of semantic coherence and a new metric: paraphrase coherence (P-coherence). In Section 4, we describe the structure of the P-thought model. In Section 5, experimental settings are described for each task, followed by results and discussions. In Section 6, we conclude the present work with some discussion of future research directions.

\section{Related work}
Recent work on sentence embedding ranges from simple extensions of the word embedding vector \cite{Doc2vec,SIF,Sent2vec,Doc2vecC,wieting2015towards} to neural network models specialized for handling a sequence of words appearing in a sentence \cite{SkipThought,InferSent}. The Distributed Bag of Words version of Paragraph Vector (PV-DBOW) and Distributed Memory Model of Paragraph Vectors (PV-DM) methods, which were proposed in Doc2vec \cite{Doc2vec}, learn sentence vectors based on the same principle: maximizing the probability to predict words in the same sentence. \citet{SIF} proposed a model that computes the sentence embedding vector as a weighted average of word embedding vectors in a sentence. By re-weighting the weights of words in a sentence, the authors achieved an improved performance in textual similarity tasks, and outperformed a complex model based on recurrent neural network (RNN). Unlike in Doc2vec \cite{Doc2vec}, in Doc2vecC \cite{Doc2vecC}, the sentence embedding vector is defined as a simple average of word embedding vectors. The idea behind doc2vecC, i.e., using an average of word embedding vectors to represent the global context of the sentence, had already been proposed by \citet{huang2012improving}. In addition, doc2vecC applies a corruption mechanism that randomly removes words from a sentence and generates a sentence embedding vector with the remaining words. This simple idea significantly reduced the total amount of training time. Similar to previous methods, Sent2vec \cite{Sent2vec} defines the sentence vector as an average of word embedding vectors. However, unlike other models using word embedding vectors of single words (i.e., uni-gram), it considers n-gram vectors in addition to uni-gram vectors when training the sentence embedding model.
	
The Skip-thought model \cite{SkipThought}, which has a sequence to sequence (Seq2Seq) structure, is an extension of the Skip-gram \cite{Skipgram} model, where the basic unit for network learning is a sentence instead of a word. Similar to the Skip-gram model, which learns word embedding vectors by training the network to predict the surrounding words when the center word is given, Skip-thought is trained to encode the input sentence and generate its preceding and following sentences. By using the generated sentence vectors as the input of a simple linear model, Skip-thought exhibited an improved performance for document classification and sentiment analysis. Inspired by previous results in computer vision, where many models are pretrained based on ImageNet \cite{Imagenet}, \citet{InferSent} conducted research on whether supervised learning tasks are helpful for learning sentence embedding vectors. Through experiments, \citet{InferSent} claimed that sentence embedding vectors generated from a model that is trained based on a natural language inference (NLI) task yield a state-of-the-art performance when leveraged in other NLP tasks. In particular, they found that a model with a bi-directional Long-short term memory (LSTM) structure and max pooling trained on the Stanford Natural Language Inference (SNLI) dataset \cite{SNLI}, named InferSent, exhibited the best performance.
	
\section{Semantic coherence}
\subsection{Defining semantic coherence}
Although two sentences may employ different words or different structures, people will recognize them as the same sentence as long as the implied semantic meanings are highly similar. Consider the following two sentences:
\begin{itemize}
	\item \textbf{Sentence 1:} Jang was caught by professor Kang while playing the computer game in the lab.
	\item \textbf{Sentence 2:} Professor Kang came to the lab and witnessed Jang playing the computer game.
\end{itemize}

Although these two sentences exhibit a clear difference with respect to both the sentence structure and word usage, people can immediately perceive that they convey the same meaning. Hence, a good sentence embedding approach should satisfy the property that if two sentences have different structures but convey the same meaning (i.e., paraphrase sentences), then they should have the same, or at least similar, embedding vectors. Based on this, we define semantic coherence as follows.\\

\textbf{Definition 1.} The degree of semantic coherence of a sentence embedding model is proportional to the similarity between the representation vectors of paraphrase sentences generated by the model.\\

If the representation vectors of paraphrase sentences are located close to each other in the embedding space, this implies that there is little difference between their vector values. Thus, when the representation vector value of a sentence is given, it should be possible to generate the given sentence and its paraphrase sentences. Consequently, we can derive the following hypothesis.\\

\textbf{Hypothesis  1}. If it is possible to generate an input sentence and its paraphrase sentence simultaneously from the vector value of the input sentence, then the sentence embedding model can enhance the semantic coherence.\\
	
In this study, we propose a new sentence embedding model to satisfy the above hypothesis.

\subsection{Evaluating semantic coherence: paraphrase coherence}
To evaluate semantic coherence, we should measure the densities of paraphrase sentences. This requires multiple pairs of paraphrase sentences that share the same meaning. Thus, previous metrics that simply calculate the matching degree of two sentences are insufficient.
	
In this study, inspired by topic coherence, which is used to determine the optimal number of topics in topic modeling, we propose a new evaluation metric called paraphrase coherence (P-coherence) to measure the semantic coherence. Topic coherence measures how effectively the highly weighted top $k$ words of a topic satisfy coherence \cite{newman2010automatic}; topic coherence is computed as follows:
\begin{gather}
Topic-coherence = 	\sum_{i<j} Score(w_i,w_j),
\end{gather}
where $w_i$ and $w_j$ are the top $i^{th}$ and $j^{th}$ words of the same topic, respectively. Although various methods exist to define the score between two words \cite{roder2015exploring}, we adopted the idea of the pointwise mutual information (PMI) measure \cite{newman2010automatic}, defined as follows:
\begin{gather}
Score_{PMI}(w_i,w_j) = \log{\frac{p(w_i,w_j)}{p(w_i)p(w_j)}},
\end{gather}
where $p(w_i,w_j)$ is the probability of words $w_i$ and $w_j$ appearing together in a randomly selected document, and $p(w_i)$ and $p(w_j)$ are the marginal probabilities that the words $w_i$ and $w_j$ appear in the randomly selected document, respectively.

Unlike in topic coherence, which defines the probability of two words appearing together based on a simple word count, we should consider the relationship between two sentences by leveraging their representation vector values. Hence, we replace the co-occurrence probability $p(w_i,w_j)$ in topic coherence with the dot product of two sentence representation vectors because the dot product of two vectors is widely used as an unnormalized probability in many studies \cite{karpathy2014deep,karpathy2015deep}. Next, we replace the marginal probability for word occurrence in topic coherence with the $L_2$-norm of the sentence vector, derived from the dot product. As a result, the score between two sentences takes a value between 0 and 1: the higher the score value, the stronger is the relationship between the two sentences. The equation representing the proposed score is as follows:
\begin{gather}
Score(sv_i,sv_j) = \frac{sv_i \cdot sv_j}{\left\|sv_i\right\|\left\|sv_j\right\|},
\end{gather}
where $sv_i$ and $sv_j$ are the representation vectors of the sentences $i$ and $j$, respectively. Finally, P-coherence is defined as the average score of all pairs of paraphrase sentences:
\begin{equation}
\begin{gathered}
P-coherence(U_k) = Average(\frac{sv_i \cdot sv_j}{\left\|sv_i\right\|\left\|sv_j\right\|}), \\
sv_i,sv_j \in U_k,	k=1,...,N,
\end{gathered}
\end{equation}
where $U_k$ is the $k^{th}$ set of paraphrase sentences, and $N$ is the number of paraphrase sets. For instance, if there are four paraphrase sentences for each paraphrase set, then the P-coherence for each paraphrase set is calculated as the average score of $_{4} C_{2}=6$ sentence pairs. The total P-coherence is the average P-coherence for each paraphrase set:
\begin{gather}
P-coherence_{Total} = \frac{1}{N} \sum_{k=1}^{N} P-coherence(U_k).
\end{gather}

\section{Paraphrase thought}
  
\subsection{Model structure}
Assume that a sentence tuple $(s,p)$ is given, where $p$ is the paraphrase sentence of the sentence $s$. Let $x_t$ be the $t^{th}$ word of the sentence $s$ and $y_t$ be the $t^{th}$ word of the sentence $p$. To maximize the semantic coherence defined above, it should be possible to generate both the sentence itself and its paraphrase sentence from the representation vector of an input sentence. Therefore, the proposed P-thought model is designed as a dual generation model, which generates both $s$ and $p$ simultaneously when the sentence tuple $(s,p)$ is given.
  
We employed an Seq2Seq structure with a gated recurrent unit (GRU) \cite{GRU} cell for the P-thought model. The $encoder$ transforms the sequence of words of an input sentence into a fixed-sized representation vector, whereas the $decoder$ generates the target sentence based on the given sentence representation vector. The proposed P-thought model has two decoders. When the input sentence is given, the first decoder, named $auto-decoder$, generates the input sentence as it is. The second decoder, named $paraphrase-decoder$, generates the paraphrase sentence of the input sentence. 
		
\subsection{Objective}
Similar to other sequence learning tasks in NLP, the purpose of the P-thought model is to minimize the negative log likelihoods of the two decoders. Furthermore, according to Hypothesis 1, the P-thought model should satisfy the condition that it can encode the sentence $s$ and generate the sentences $s$ and $p$ simultaneously when the sentence pair $(s,p)$ is given. This condition can be written as follows:
\begin{gather}
P(s)P(s|s;\theta_{ss}) = P(s)P(p|s;\theta_{sp}),
\end{gather}
where $P(s)$ is the marginal probability of the input sentence $s$, and $\theta_{ss}$ and $\theta_{sp}$ are the parameters of $auto-decoder$ and $paraphrase-decoder$, respectively. Thus, similarly to the work of \citet{DSL}, the problem can be formulated as the following multi-objective optimization problem:
\begin{equation}
\begin{gathered}
\textbf{Objective 1}: \min_{\theta_{ss}}l_{A}(f(s;\theta_{ss}),s),\\
\textbf{Objective 2}: \min_{\theta_{sp}}l_{P}(f(s;\theta_{sp}),p),\\
s.t \quad P(s)P(s|s;\theta_{ss}) = P(s)P(p|s;\theta_{sp}),
\end{gathered}
\end{equation}
where $l_A$ and $l_P$ are the negative log likelihoods of $auto-decoder$ and $paraphrase-decoder$, respectively. In this case, the constraint term can be rewritten as follows:
\begin{gather}
-\log{P(s|s;\theta_{ss})} = -\log{P(p|s;\theta_{sp})},
\end{gather}
The left and right terms of transformed equation represent the negative log likelihoods of $auto-decoder$ and $paraphrase-decoder$, respectively. Hence, the constraint term can be written as follows:
\begin{gather}
l_{A}(f(s;\theta_{ss}),s) = l_{P}(f(s;\theta_{sp}),p).
\end{gather}
By introducing the Lagrange multiplier, the multi-objective optimization problem is transformed into the following minimization problem:
\begin{align}
\min L = & l_{A}(f(s;\theta_{ss}),s) + l_{P}(f(s;\theta_{sp}),p) \nonumber \\
& - \lambda(l_{A}(f(s;\theta_{ss}),s) - l_{P}(f(s;\theta_{sp}),p)) \nonumber \\
= & (1-\lambda)l_{A}(f(s;\theta_{ss}),s) + (1+\lambda)l_{P}(f(s;\theta_{sp}),p)), \nonumber \\
& where \quad \lambda \neq 0.
\end{align}
In this case, a value of $\lambda>1$ or $\lambda<-1$ leads to maximizing the negative log likelihood of $auto-decoder$ and that of $paraphrase-decoder$, respectively. To avoid this problem, the allowable range for $\lambda$ is set to $-1<\lambda<0$ or $0<\lambda<1$. Under this condition, minimizing $L$ is equivalent to minimizing $L^\prime$:
\begin{equation}
\begin{gathered}
\min L^\prime = l_{A}(f(s;\theta_{ss}),s) + \alpha l_{P}(f(s;\theta_{sp}),p),\\
where \quad \alpha = \frac{1+\lambda}{1-\lambda},
\end{gathered}
\end{equation}
where $\alpha$ is the hyperparameter of the P-thought model. This should be greater than 1 or be the value between 0 and 1, because $-1<\lambda<0$ and $0<\lambda<1$. However, it is desirable to set the appropriate $\alpha$-value to greater than 1, considering that auto decoding is trivial copying task which is much easier than paraphrase generation. Experimental results also demonstrated that the performance is degraded for small $\alpha$ value. Thus, the objective of the P-thought model is the sum of the negative log likelihood of $auto-decoder$ with that of $paraphrase-decoder$ with a higher weight:
\begin{equation}
\begin{gathered}
Loss = l_{auto}(f(s;\theta_{ss}),s) + \alpha l_{para}(f(s;\theta_{sp}),p),\\
where \quad \alpha > 1.
\end{gathered}
\end{equation}

\begin{figure}[t!]%
	\centering
	\begin{subfigure}[b]{0.34\textwidth}
		\includegraphics[width=\linewidth]{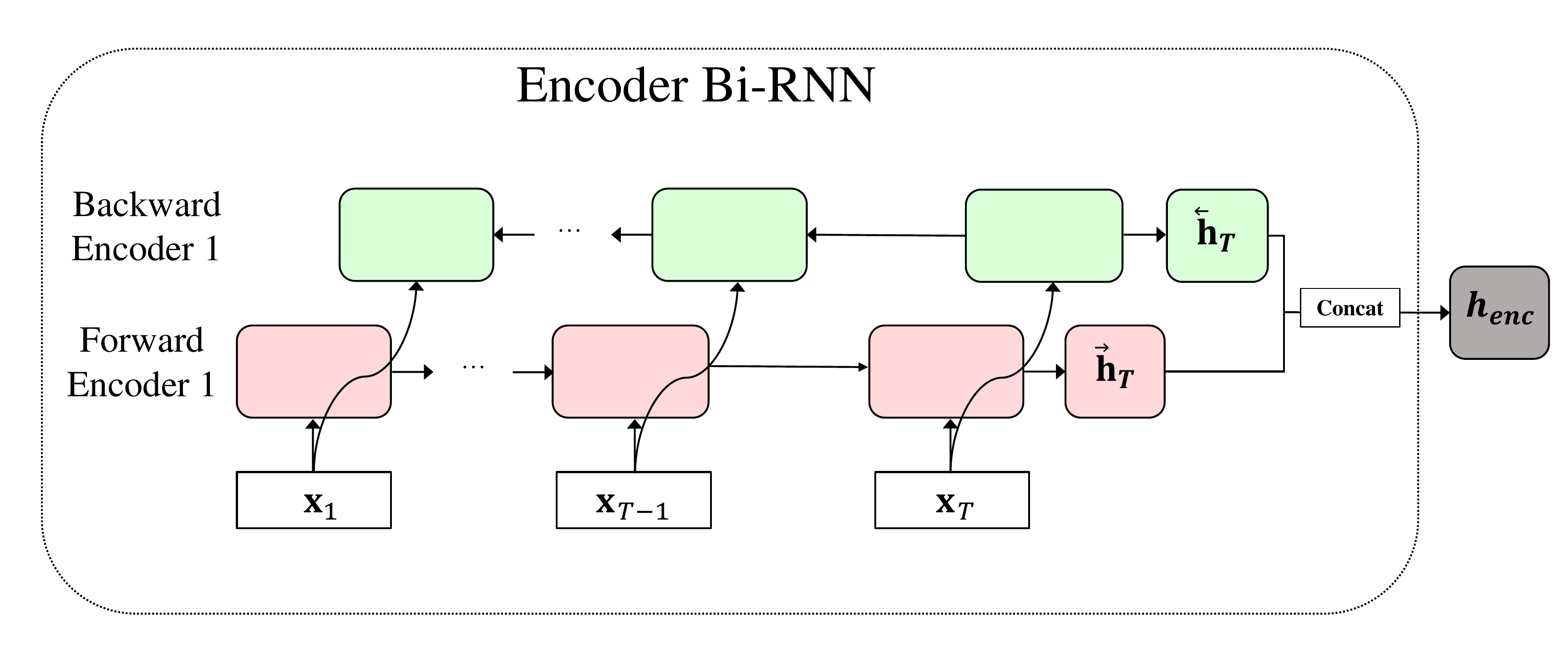}
		\caption{One-layer Bi-RNN encoder}
		\label{fig:Enc_lv1}
	\end{subfigure}%
	\hfil
	\begin{subfigure}[b]{0.34\textwidth}
		\includegraphics[width=\linewidth]{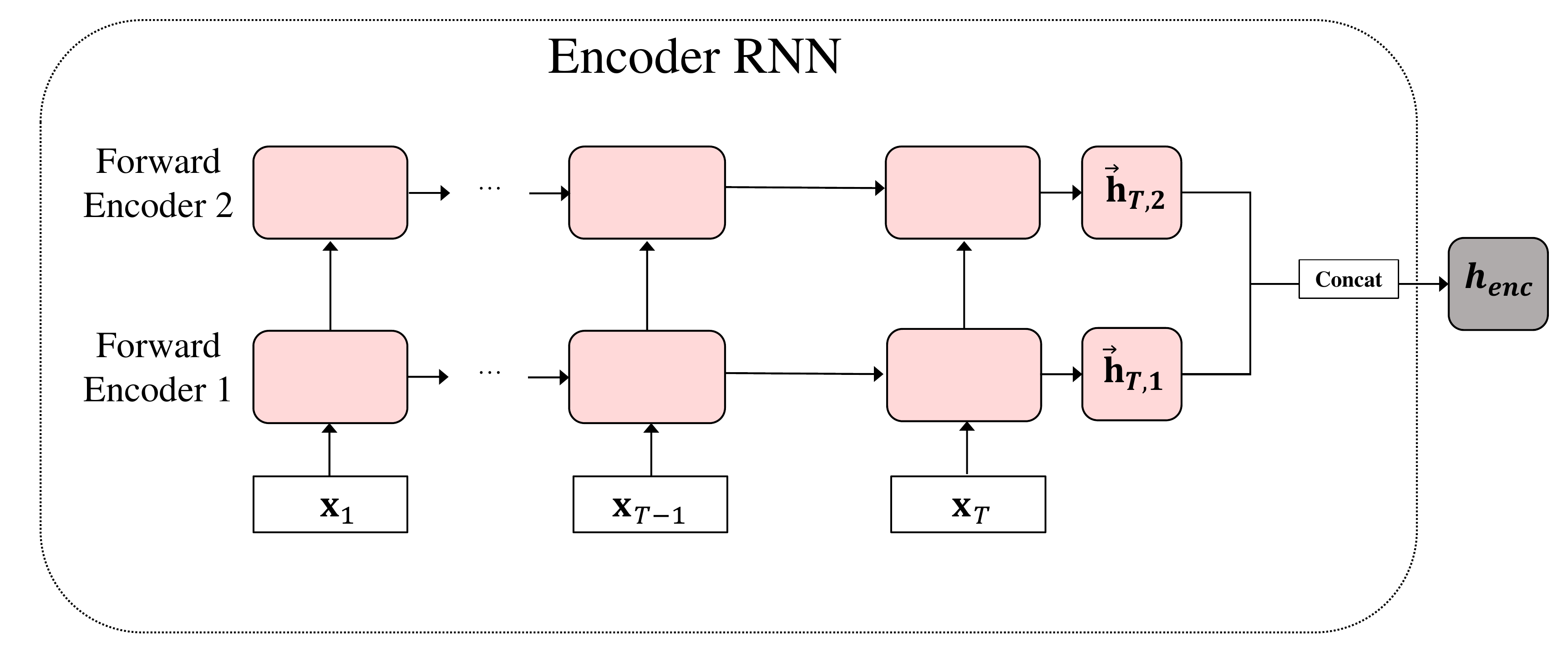}
		\caption{Two-layer forward RNN}
		\label{fig:Enc_lv2}
	\end{subfigure}%
	\hfil
	\begin{subfigure}[b]{0.34\textwidth}
		\includegraphics[width=\linewidth]{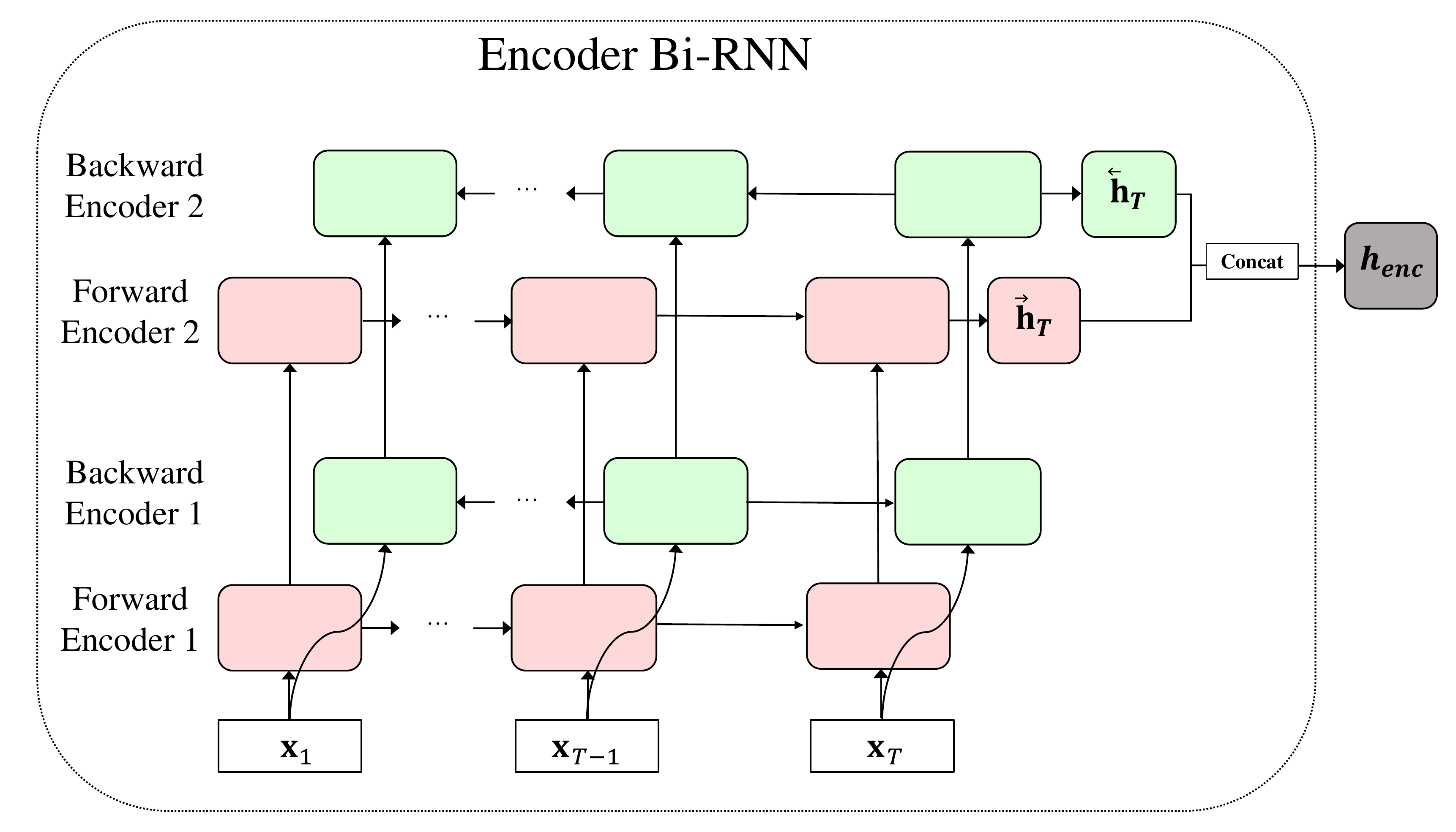}
		\caption{Two-layer BiRNN}
		\label{fig:Enc_lv3}
	\end{subfigure}%
	\caption{Three encoder structures of the P-thought model}\label{fig:encoder_structure}
\end{figure}

\subsection{Vocabulary expansion}
The number of unique words appearing in our training dataset is only about 35,000, which is considerably fewer than the number of words in the English language. This may be problematic, in that many words are treated as out of vocabulary after model training. To solve this problem, motivated by the idea of cross-lingual embedding \cite{mikolov2013exploiting}, Skip-thought attempts to learn a matrix that maps the words of a pretrained word2vec model \cite{Skipgram} to one of 20,000 words in their training dataset. However, this approach suffers from the problem that a word can be mapped to another word whose actual meaning is significantly different, only because it has a high similarity with the original word in the embedding space. For example, the word 'endogenous' was mapped to the word 'neuronal,' despite the semantic differences.

We extracted the vector values of words that appear in our training dataset from the pretrained Glove vector \cite{Glove} to resolve the problem described above. In the pretrained Glove vectors, the semantic relationships between words are reflected in the geometrical structures between word vectors. Therefore, even when vector values of words that are unused during training occur, the information loss can be reduced because the geometric relationships between word vectors are well preserved if the model is sufficiently trained. By using this method, we are able to handle 2.1 million words without the effort of training an extra mapping matrix.

\begin{figure}[t!]
	\centering
	\begin{subfigure}[b]{0.45\textwidth}
		\includegraphics[width=\linewidth]{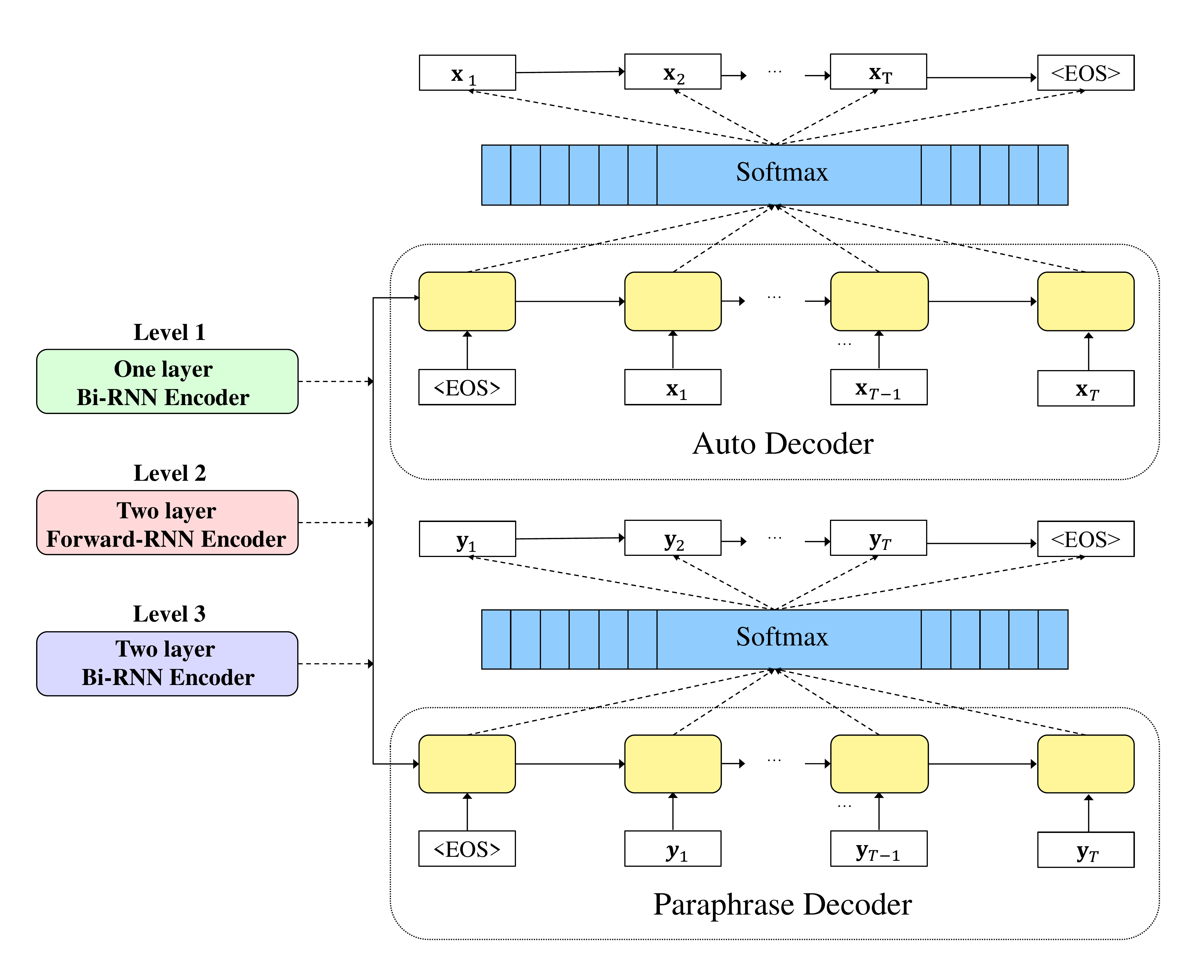}
	\end{subfigure}%
	\caption{P-thought model structure}
	\label{fig:decoder}
\end{figure}

\section{Experiments}
\subsection{Experimental settings}
We used the captions of the MS-COCO dataset \cite{MSCOCO} to train the P-thought model. This dataset has been employed in various paraphrase generation studies \cite{prakash2016neural,gupta2017deep}. The MS-COCO dataset has more than five captions for each image, which allows us to generate more than $_{5} P_{2}=20$ unique sentence pairs. For training, we used the 2014-Validation and 2017-Training datasets. Descriptions of these datasets are provided in Table \ref{table1.Training data description}. Simple tokenizing was performed as text preprocessing for the captions.

\begin{table}[t!]
	\begin{center}
		\caption{Training data description} \label{table1.Training data description}%
		\renewcommand{\arraystretch}{1}
		\footnotesize{
		\centering{\setlength\tabcolsep{4pt}
			\begin{tabular}{c|c|c|c}
				\hline
				& 2014-Validation & 2017-Training & Total \\
				\hline
				No. of unique & \multirow{2}{*}{40,504} & \multirow{2}{*}{118,284} & \multirow{2}{*}{123,287} \\
				images & & & \\
				\hline
				No. of unique & \multirow{2}{*}{202,654} & \multirow{2}{*}{591,753} & \multirow{2}{*}{593,968} \\
				captions & & & \\
				\hline
				No. of unique & \multirow{2}{*}{811,426} & \multirow{2}{*}{2,368,926} & \multirow{2}{*}{2,467,293} \\
				sentence pairs & & & \\
				\hline
				No. of unique & \multirow{2}{*}{-} & \multirow{2}{*}{-} & \multirow{2}{*}{34,826} \\
				words & & & \\
				\hline
		\end{tabular}}}
	\end{center}
\end{table}

We employed three different encoder structures, as shown in Figure \ref{fig:encoder_structure}, to investigate the model performances according to different levels of model complexity. The first encoder structure has one layer with a bi-directional RNN (Bi-RNN). The sentence embedding vector of this encoder structure consists of the concatenated values of the final state values of the forward and backward RNN. The second encoder structure contains two layers, with only a forward RNN. The sentence embedding vector is generated by concatenating the final states of both layers. The third encoder structure contains two layers of Bi-RNN. The sentence embedding vector is generated from the concatenated values of the final states of the second layer's forward and backward RNNs. The overall structure of the P-thought model, including the decoder part, is illustrated in Figure \ref{fig:decoder}. These three models were trained under the same conditions. The number of hidden units is set to 1,200, which results in 2,400-dimensional sentence embedding vectors after concatenation. We employed Xavier initialization \cite{Xavier}, and gradient computations and weight updates were performed with a mini-batch size of 128. All models were trained for four epochs using the Adam optimizer \cite{Adam}.

\begin{table}[t!]
	 	\begin{center}
	 		\caption{2017-Validation dataset description} \label{table2.Testing data description}%
	 		\renewcommand{\arraystretch}{1}
	 		\footnotesize{
	 			\centering{\setlength\tabcolsep{1pt}
	 				\begin{tabular}{c|c|c|c}
	 					\hline
	 					\makecell{No. of unique\\images} & \makecell{No. of unique\\captions} & \makecell{No. of unique \\sentence pair} & \makecell{No. of unique \\words} \\ 
	 					\hline
	 					5,000 &  25,014 &  100,142 &  8,641 \\ 
	 					\hline
	 		\end{tabular}}}
	 	\end{center}
	 \end{table}
	 
\begin{table}[t!]
	\begin{center}
		\caption{Experimental results for evaluating the P-coherence} \label{table3.P-coherence result}%
		\renewcommand{\arraystretch}{1}
		\footnotesize{
		\centering{\setlength\tabcolsep{10pt}
			\begin{tabular}{c|c}
				\hline
				\textbf{Model} & \textbf{P-coherence} \\ 
				\hline
				PV-DBOW &  \multirow{2}{*}{0.0099} \\ 
				\cite{Doc2vec}  & \\
				\hline
				Uni-skip &  \multirow{2}{*}{0.5328} \\
				\cite{SkipThought} & \\ 
				\hline
				Bi-skip  &  \multirow{2}{*}{0.5155} \\ 
				\cite{SkipThought} & \\
				\hline
				Combine-skip &  \multirow{2}{*}{0.5209} \\ 
				\cite{SkipThought} & \\
				\hline
				SIF &  \multirow{2}{*}{0.4205} \\ 
				\cite{SIF} & \\
				\hline
				Sent2vec Wiki-uni &  \multirow{2}{*}{0.4279} \\ 
				\cite{Sent2vec}  & \\
				\hline
				Sent2vec Wiki-bi  &  \multirow{2}{*}{0.4553} \\ 
				\cite{Sent2vec} & \\
				\hline
				InferSent &  \multirow{2}{*}{0.7454} \\ 
				\cite{InferSent} & \\ 
				\hline
				P-thought &  \multirow{2}{*}{0.7432} \\ 
				(one layer-Bi RNN) & \\
				\hline
				P-thought &  \multirow{2}{*}{\textbf{0.7899}} \\ 
				(two layers-Forward RNN) & \\
				\hline
				P-thought &  \multirow{2}{*}{\textbf{0.9725}} \\ 
				(two layers-Bi RNN) & \\
				\hline
		\end{tabular}}}
	\end{center}
\end{table}

 \begin{figure*}
 	\centering
 	\begin{subfigure}[b]{0.25\textwidth}
 		\includegraphics[width=\linewidth]{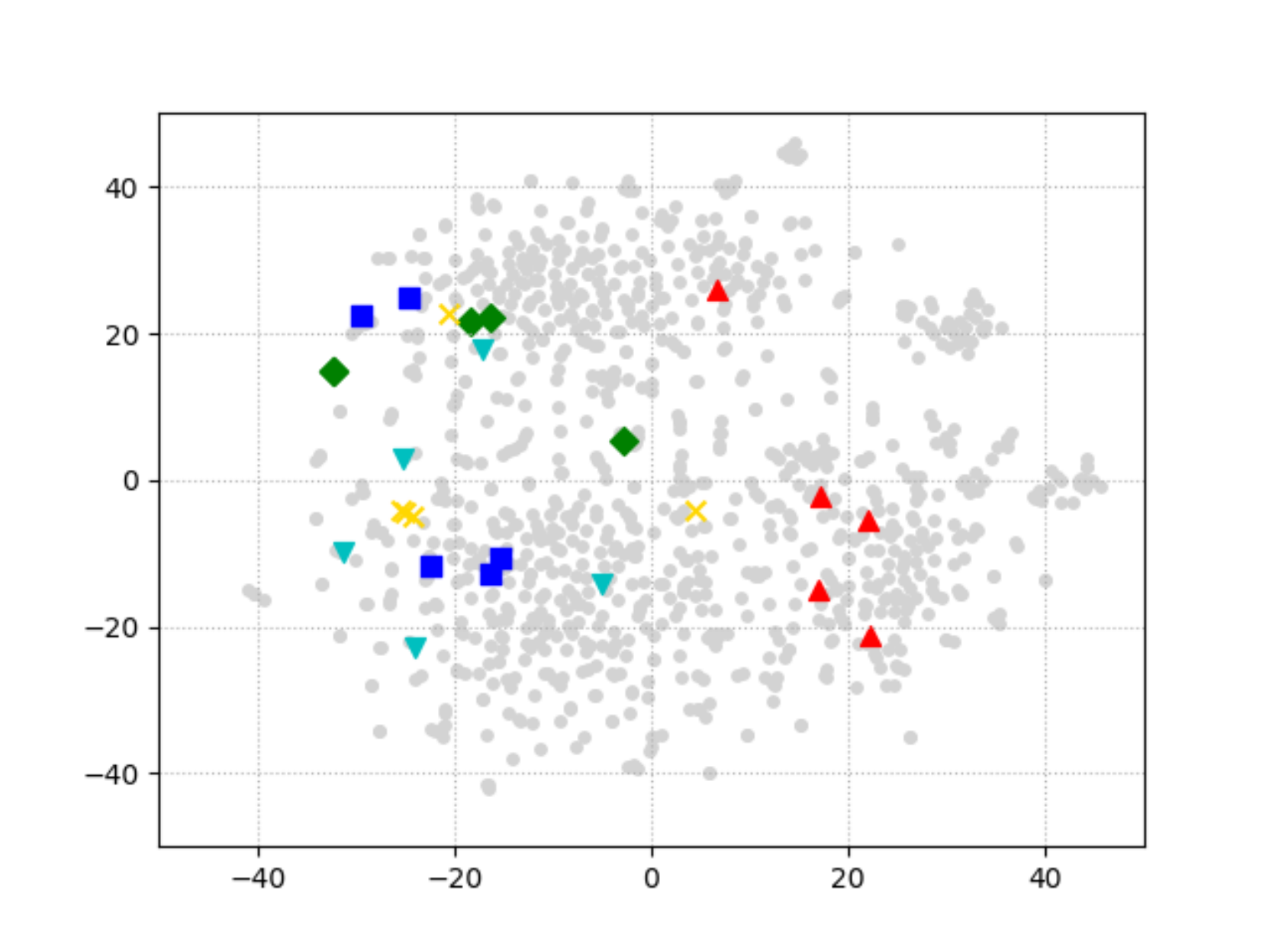}
 		\caption{Uni-skip}
 		\label{fig:uni-skip}
 	\end{subfigure}%
 	\begin{subfigure}[b]{0.25\textwidth}
 		\includegraphics[width=\linewidth]{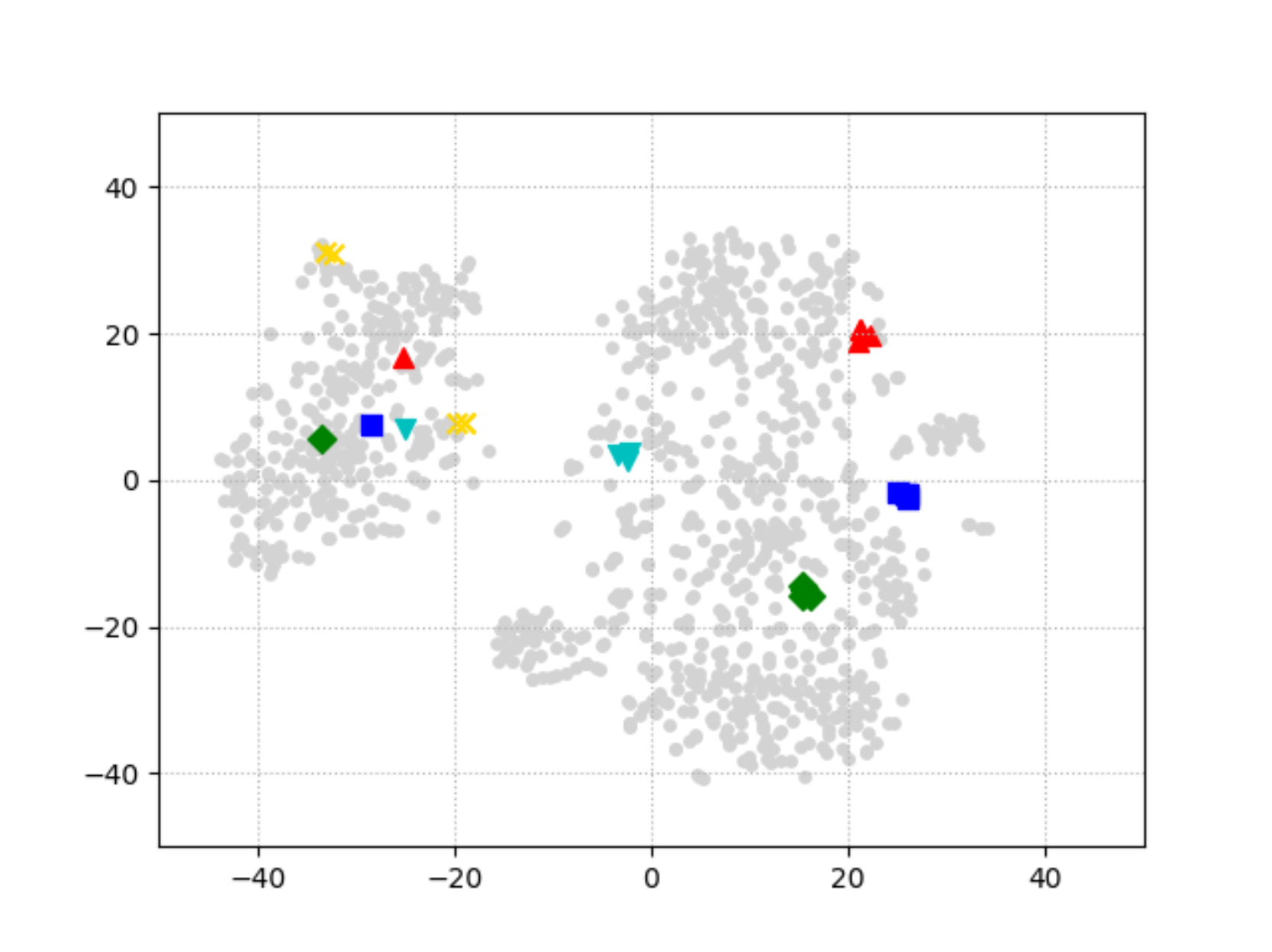}
 		\caption{SIF}
 		\label{fig:SIF}
 	\end{subfigure}%
 	\begin{subfigure}[b]{0.25\textwidth}
 		\includegraphics[width=\linewidth]{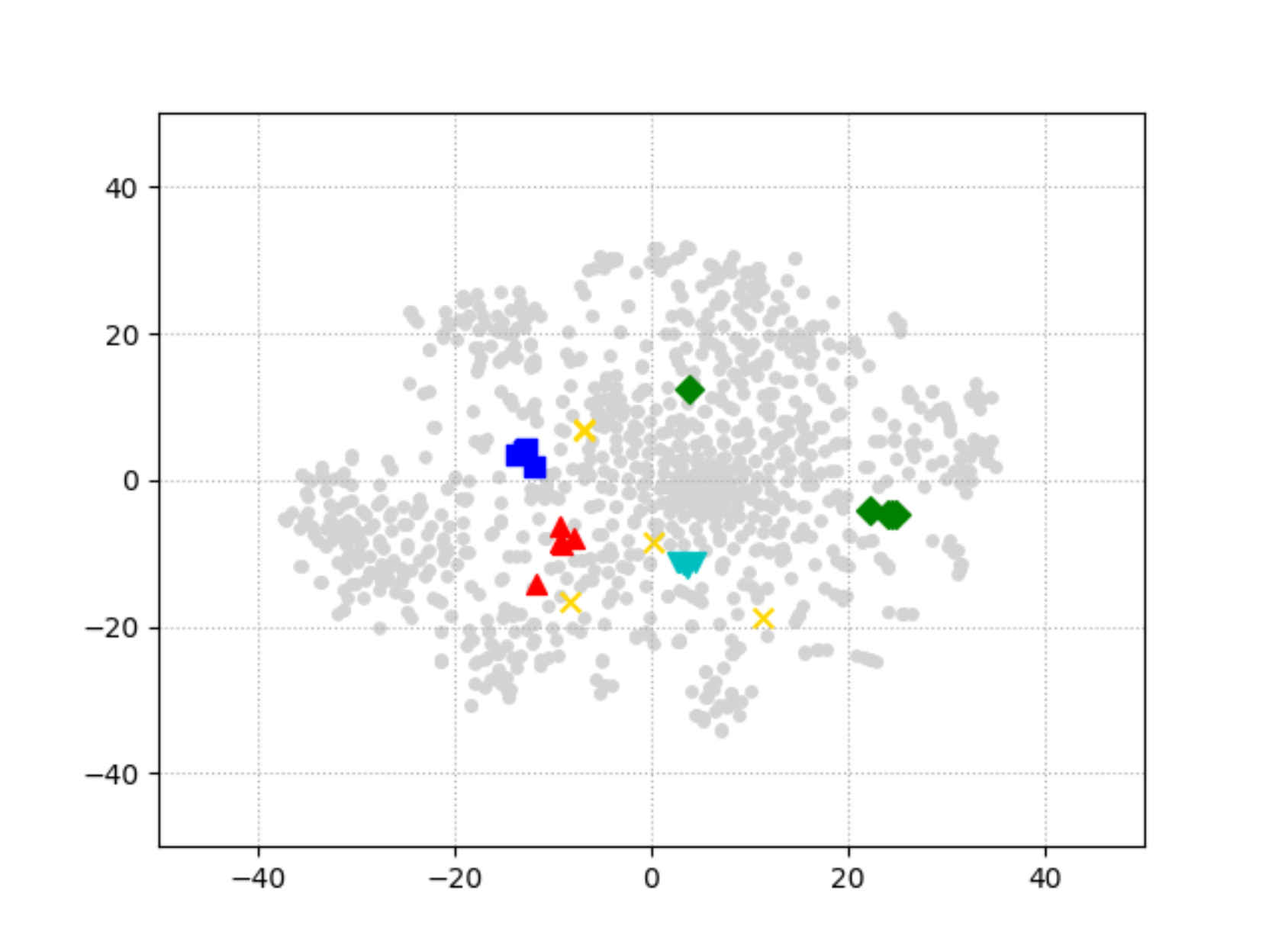}
 		\caption{Sent2vec (Wiki-uni)}
 		\label{fig:Sent2vec_uni}
 	\end{subfigure}%
 	\begin{subfigure}[b]{0.25\textwidth}
 		\includegraphics[width=\linewidth]{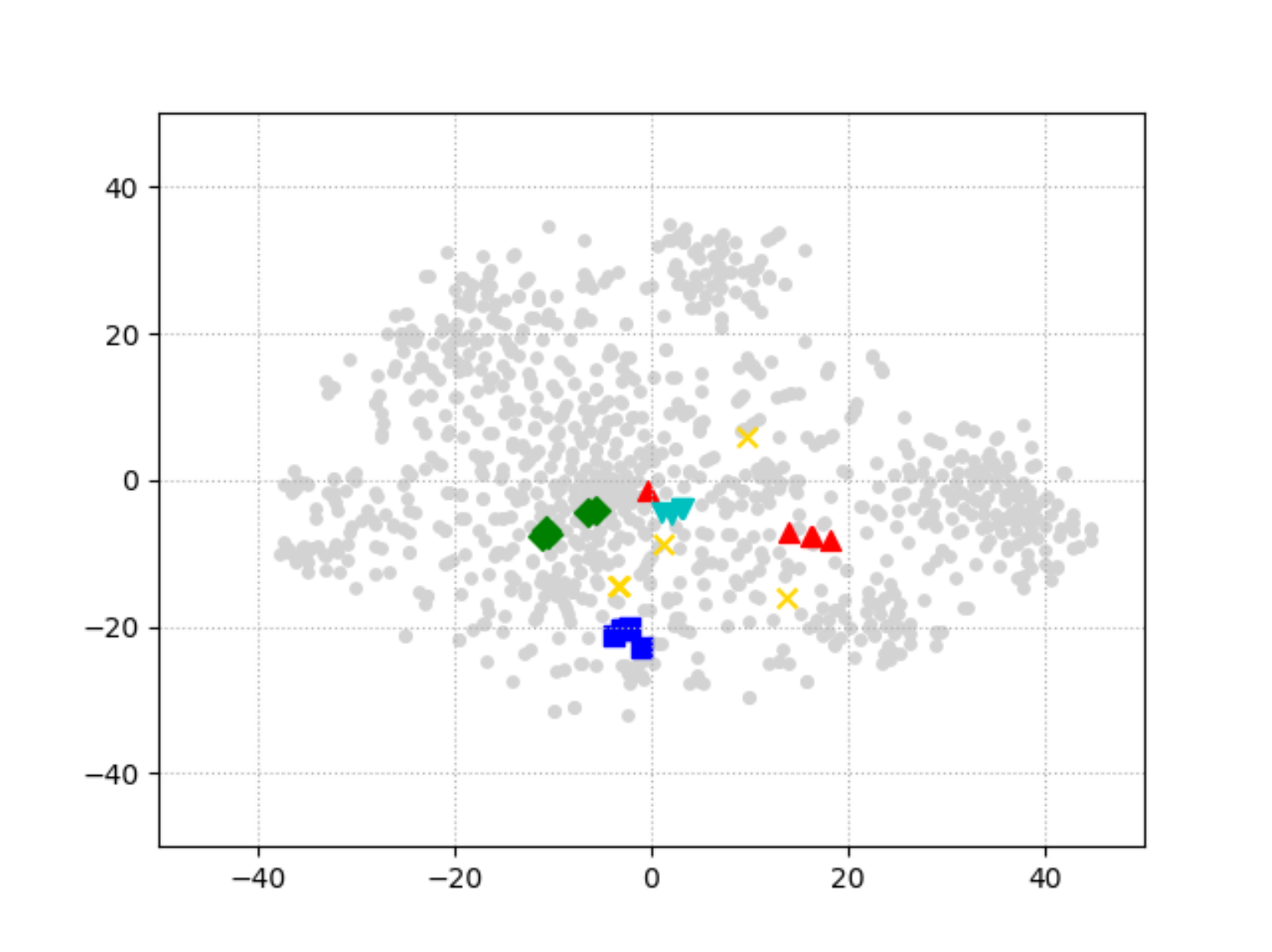}
 		\caption{Sent2vec (Wiki-bi)}
 		\label{fig:Sen2vec_bi}
 	\end{subfigure}
 	\hfil
	\begin{subfigure}[b]{0.25\textwidth}
	\includegraphics[width=\linewidth]{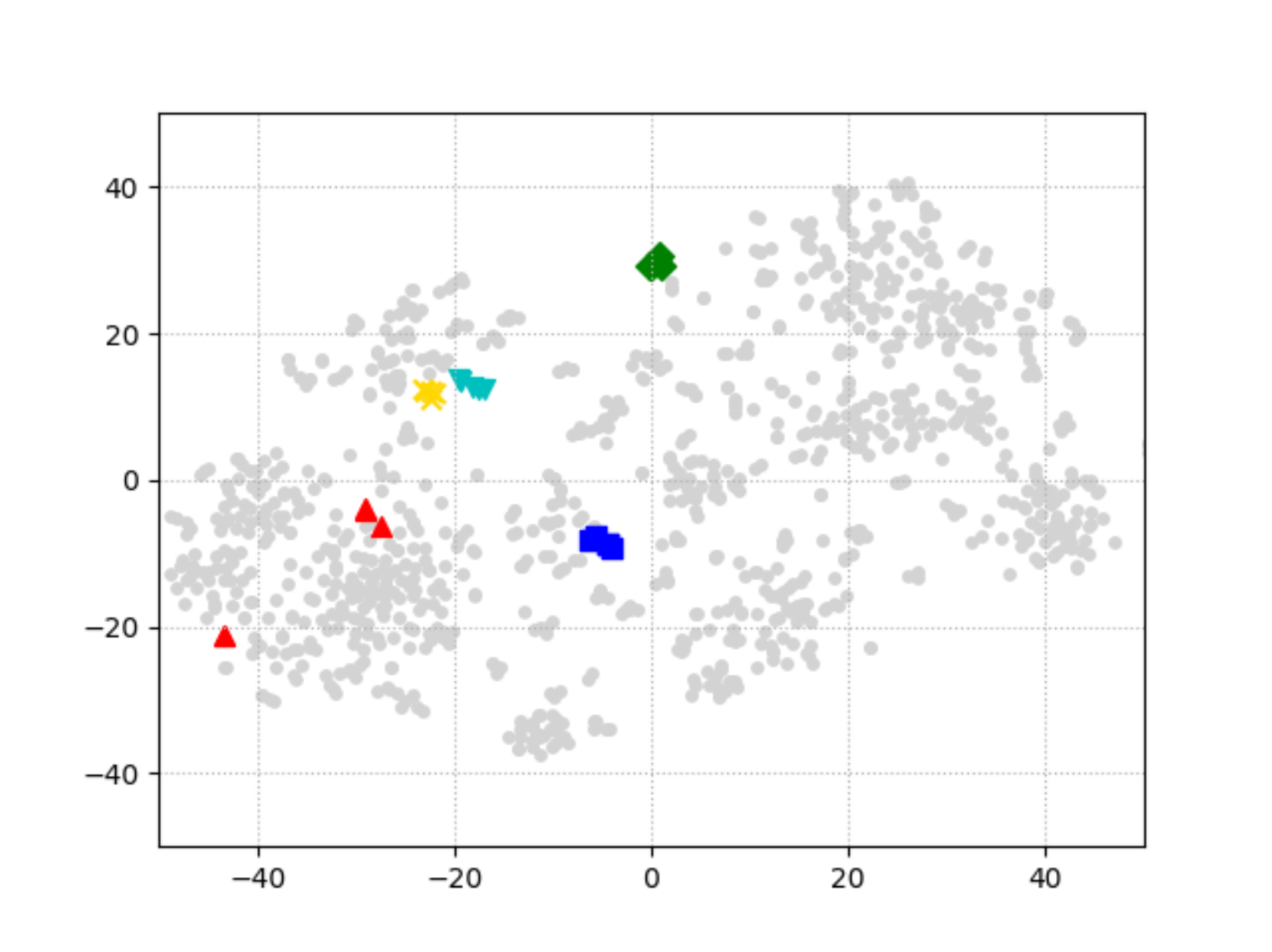}
	\caption{InferSent}
	\label{fig:InferSent}
	\end{subfigure}%
	\begin{subfigure}[b]{0.25\textwidth}
	\includegraphics[width=\linewidth]{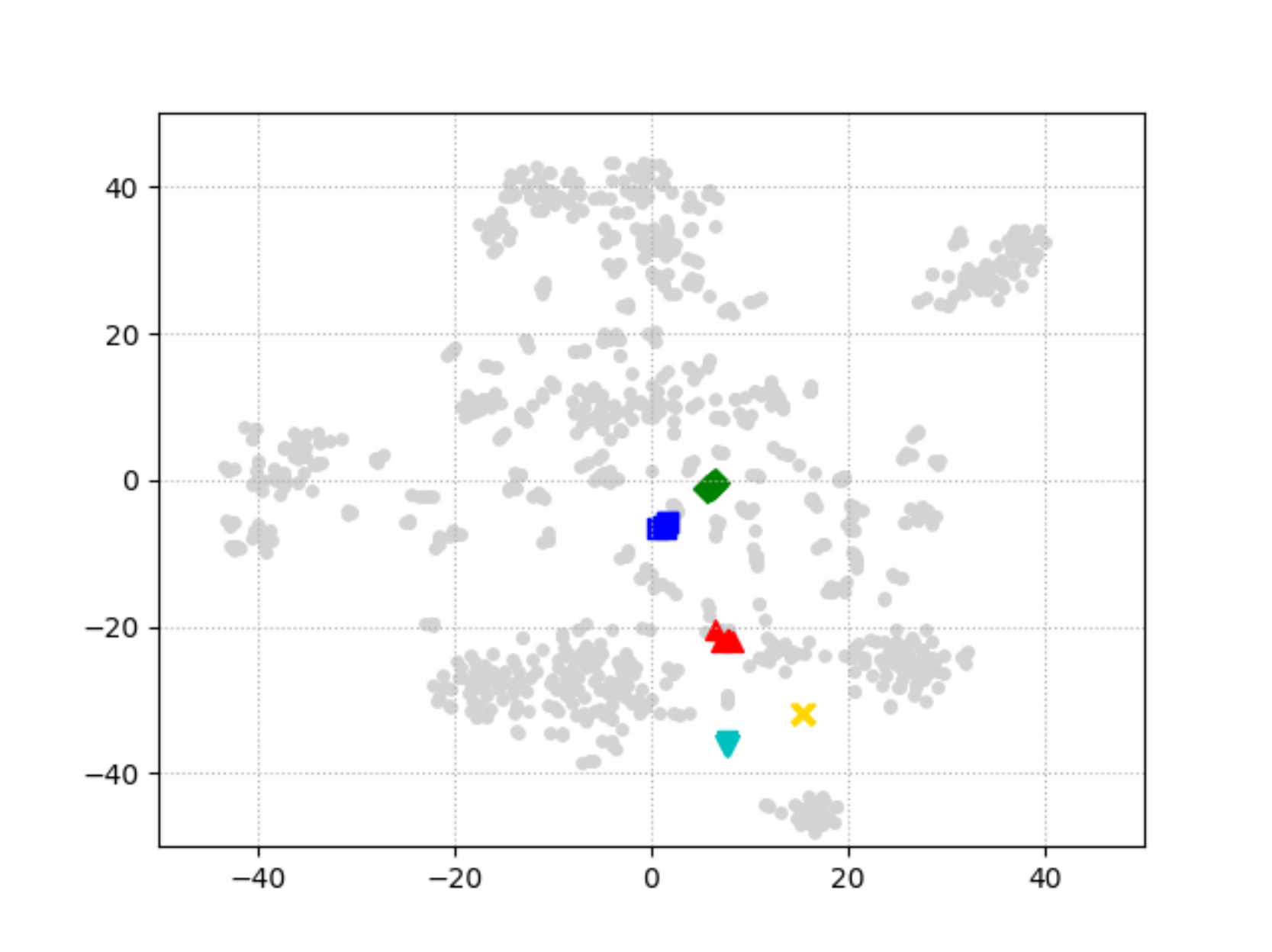}
	\caption{P-thought (One-Bi)}
	\label{fig:one_bi}
	\end{subfigure}%
	\begin{subfigure}[b]{0.25\textwidth}
	\includegraphics[width=\linewidth]{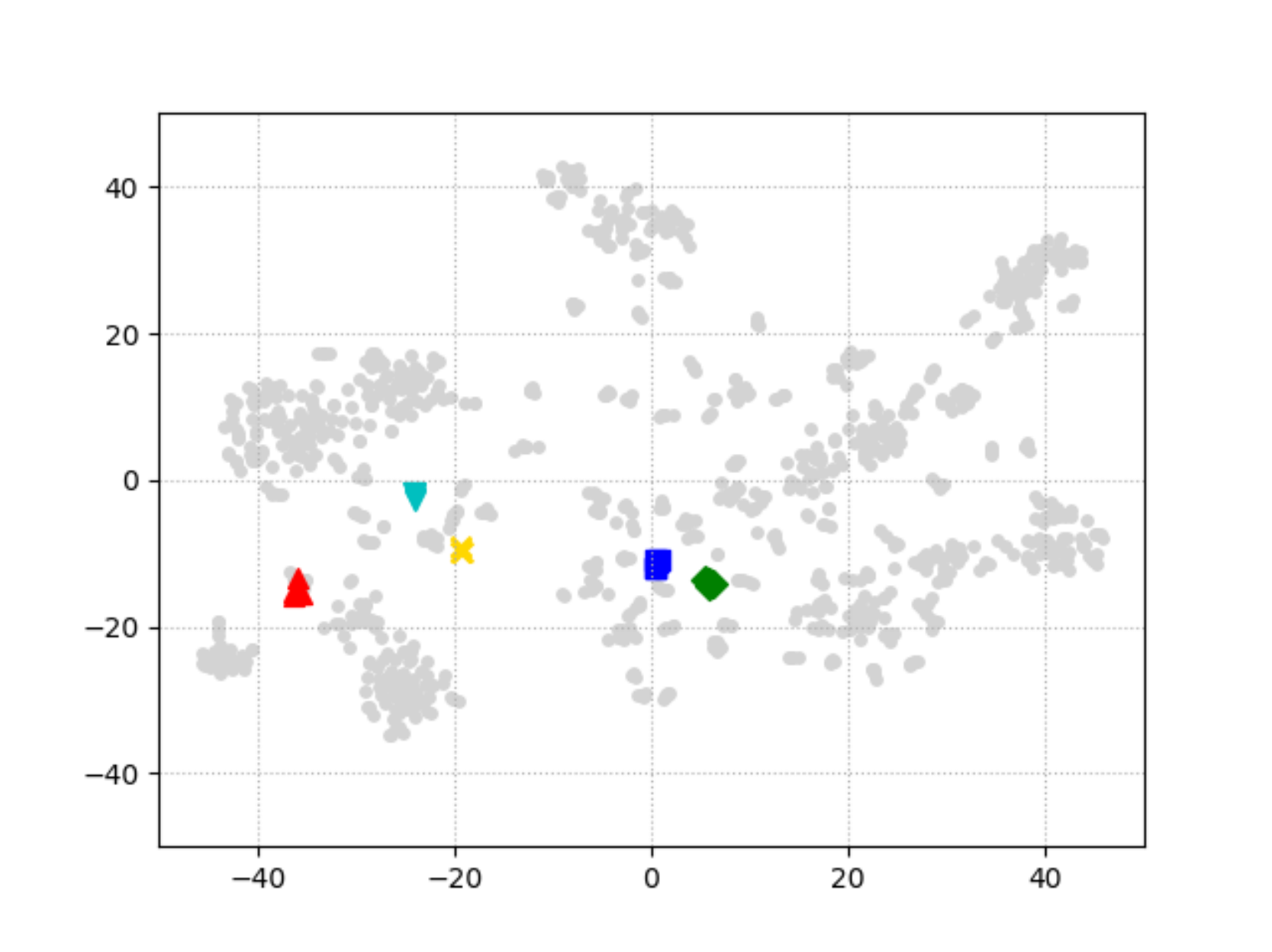}
	\caption{P-thought (Two-Forward)}
	\label{fig:two_forward}
	\end{subfigure}%
	\begin{subfigure}[b]{0.25\textwidth}
	\includegraphics[width=\linewidth]{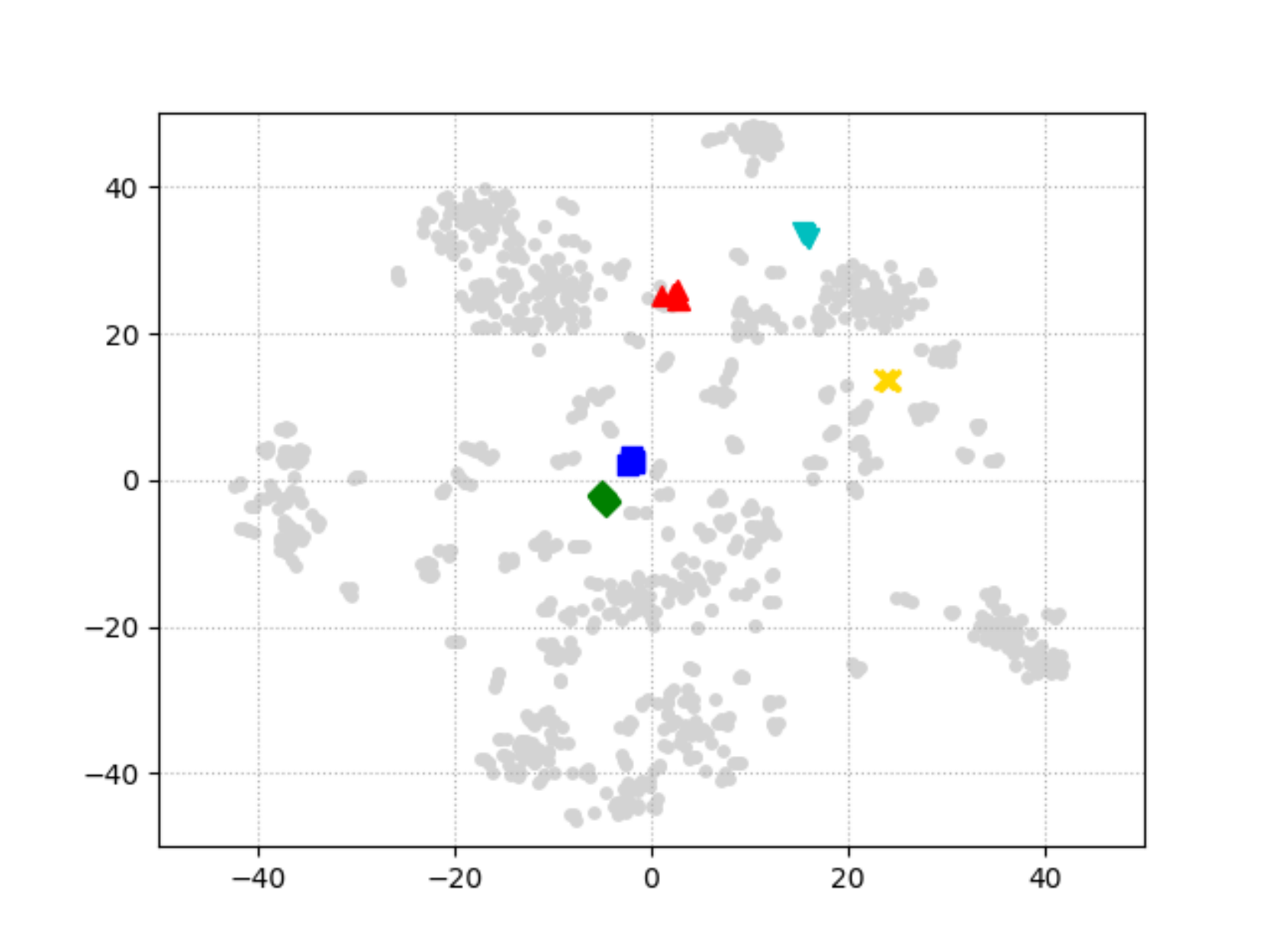}
	\caption{P-thought (Two-Bi)}
	\label{fig:two_bi}
	\end{subfigure}
	\caption{Scatter plots of the five paraphrase sentence groups represented by each sentence embedding method}\label{fig:Scatterplot}
 	\end{figure*}
 
 \begin{table}[t!]
 	\begin{center}
 		\caption{Extracted sentences for visualization} \label{table4.sample sentences}%
 		\small{
 			\centering{\setlength\tabcolsep{2pt}
 				\begin{tabular}{m{1\linewidth}}
 					\hline
 					\textbf{\textcolor{red}{Group 1 ($\blacktriangle$)}} \\
 					1) Bedroom scene with a bookcase, blue comforter and window.\\ 
 					2) A bedroom with a bookshelf full of books.\\
 					3) This room has a bed with blue sheets and a large bookcase. \\	
 					4) A bed and a mirror in a small room.\\
 					5) A bed room with a neatly made bed a window and a book shelf \\ 
 					\hline
 					\textbf{\textcolor[RGB]{0,100,0}{Group 2 ($\blacklozenge$)}} \\ 
 					1) A male tennis player in white shorts is playing tennis. \\ 
 					2) This woman has just returned a volley in tennis. \\
 					3) A man holding a tennis racket playing tennis. \\	
 					4) The man balances on one leg after serving a tennis ball. \\
 					5) Someone playing in a tennis tournament with a crowd looking on. \\ 
 					\hline
 					\textbf{\textcolor{blue}{Group 3 ($\blacksquare$)}} \\
 					1) A woman holding a Hello Kitty phone on her hand. \\ 
 					2) A woman holds up her phone in front of her face. \\
 					3) A woman in white shirt holding up a cellphone. \\	
 					4) A woman checking her cell phone with a hello kitty case. \\
 					5) The Asian girl is holding her Miss Kitty phone. \\ 
 					\hline
 					\textbf{\textcolor[RGB]{255,215,0}{Group 4 ($\times$)}} \\
 					1) A plate of food which includes onions, tomato, lettuce, sauce, fries, and a sandwich. \\ 
 					2) A sandwich, french fries, bowl of ketchup, onion slice, 
 					lettuce slice, tomato slice, and knife sit on the white plate. \\
 					3) Partially eaten hamburger on a plate with fries and condiments. \\	
 					4) A grilled chicken sandwich sits beside french fries made with real potatoes. \\
 					5) A sandwich on a sesame seed bun next to a pile of french fries and a cup of ketchup \\ 
 					\hline
 					\textbf{\textcolor{cyan}{Group 5 ($\blacktriangledown$)}} \\  
 					1) Decorated coffee cup and knife sitting on a patterned surface. \\ 
 					2) A large knife is sitting in front of a mug has a skull and crossbones. \\
 					3) A white mug showing pirate skull and bones and a large knife on a counter top. \\	
 					4) There is a white coffee cup with a skull and bones on it next to a knife. \\
 					5) A close up of a knife and a cup on a surface \\ 
 					\hline
 		\end{tabular}}}
 	\end{center}
 \end{table}

\subsection{P-coherence}
To measure the P-coherence, we used the 2017-Validation dataset from the MS-COCO caption dataset, which has no overlap with the training dataset. A description of the dataset used for evaluating the P-coherence is provided in Table \ref{table2.Testing data description}. We selected PV-DBOW, Skip-thought, SIF, Sent2vec, and InferSent as benchmark models. In the case of PV-DBOW, we employed the datasets used for both training P-thought and evaluating the P-coherence to learn the sentence vectors. For the remaining models, we used the publicly available pretrained models. 

The experimental results are summarized in Table \ref{table3.P-coherence result}. It can be observed that the P-thought models with relatively complex encoder structures outperformed other benchmarked models. In the case of P-thought with a one-layer Bi-RNN, the P-coherence value is comparable to that of InferSent, and superior to the other benchmarked models. Among the benchmarked models, InferSent yielded a significantly higher P-coherence value than the other models, which implies that InferSent preserved the semantic coherence when learning the sentence representation vectors. 

In addition to the quantitative evaluation provided in Table \ref{table3.P-coherence result}, we reduced the generated sentence vectors to two-dimensional vectors using $t$-SNE \cite{T-SNE} and created scatter plots to qualitatively investigate how effectively the paraphrase sentences satisfied coherence. For the sake of visualization, we extracted the paraphrase sentences for five images and marked them with different colors and shapes. The extracted paraphrase sentences are presented in Table \ref{table4.sample sentences}, and the scatter plots are given in Figure \ref{fig:Scatterplot}. It can easily be observed that paraphrase sentence vectors learned by the models with high P-coherence values (P-thought and InferSent) are more concentrated than those of the other models.

\subsection{STS Benchmark task}

We also carried out the STS Benchmark task \cite{STS_Benchmark} to evaluate how well the models preserve the meanings of sentences through a more generally conducted task. The dataset for this task consists of 8,628 sentence pairs and corresponding human rated similarity scores valued between 0 and 5. The purpose of this task is to approximate the similarity scores between sentences based on the embedded vectors. A description of the dataset is summarized in Table \ref{table5.STS data description}.

\begin{figure}
	\begin{subfigure}[b]{0.25\textwidth}
		\includegraphics[width=\linewidth]{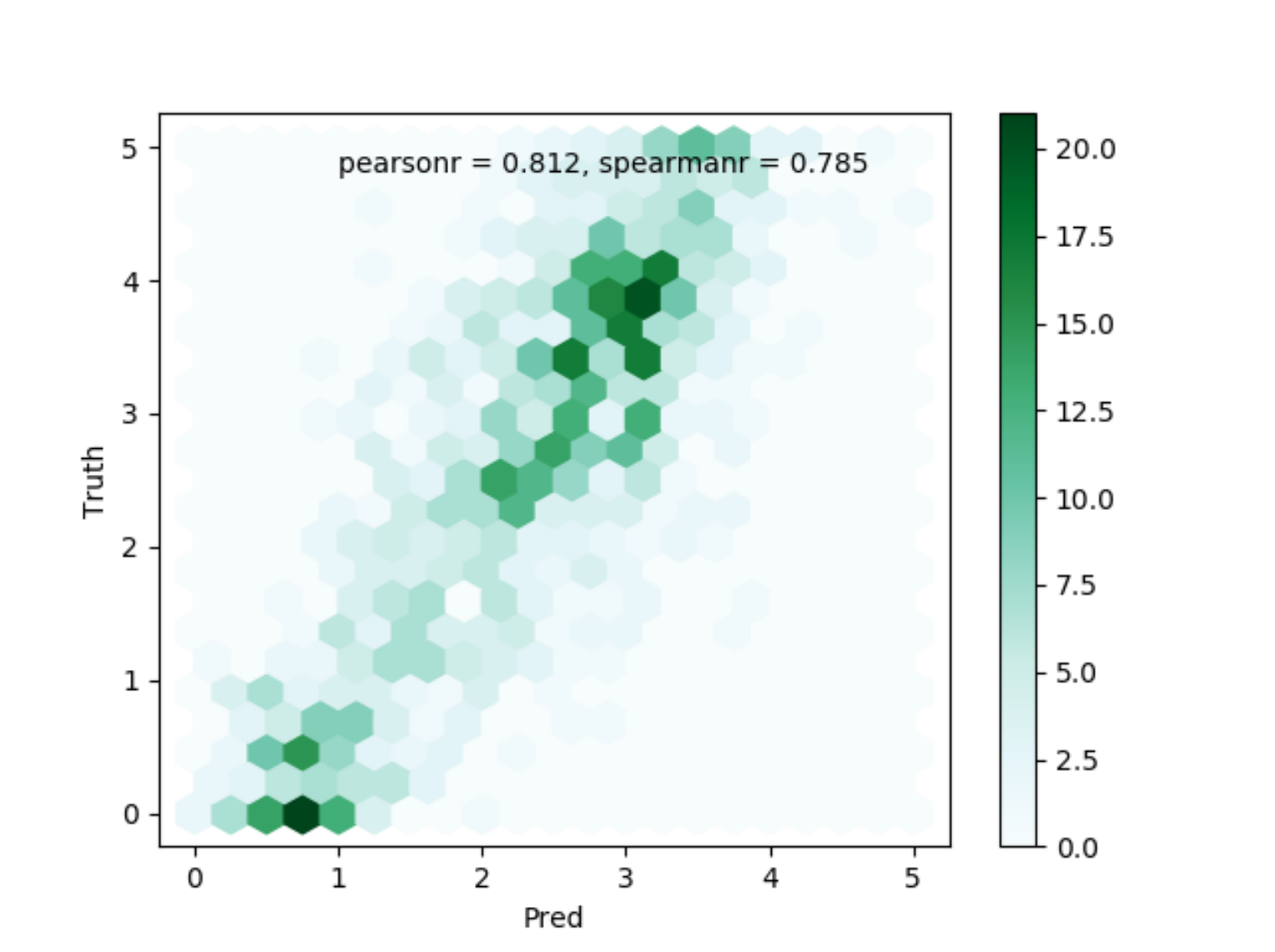}
		\caption{One-layer BiRNN}
		\label{fig:STS_onebi}
	\end{subfigure}%
	\begin{subfigure}[b]{0.25\textwidth}
		\includegraphics[width=\linewidth]{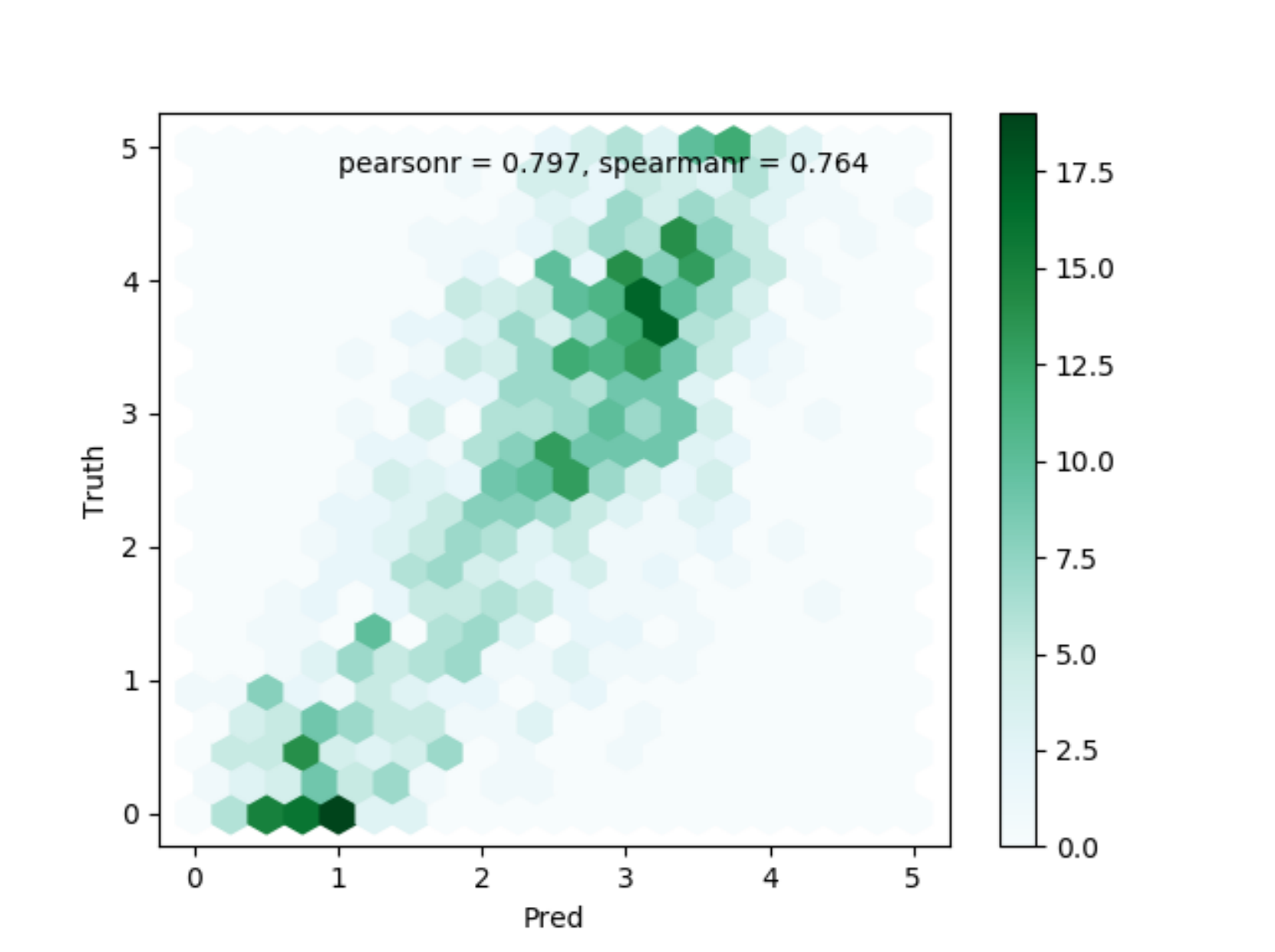}
		\caption{Two-layer forward RNN}
		\label{fig:STS_twoforward}
	\end{subfigure}%
	\hfil
	\begin{subfigure}[b]{0.25\textwidth}
		\includegraphics[width=\linewidth]{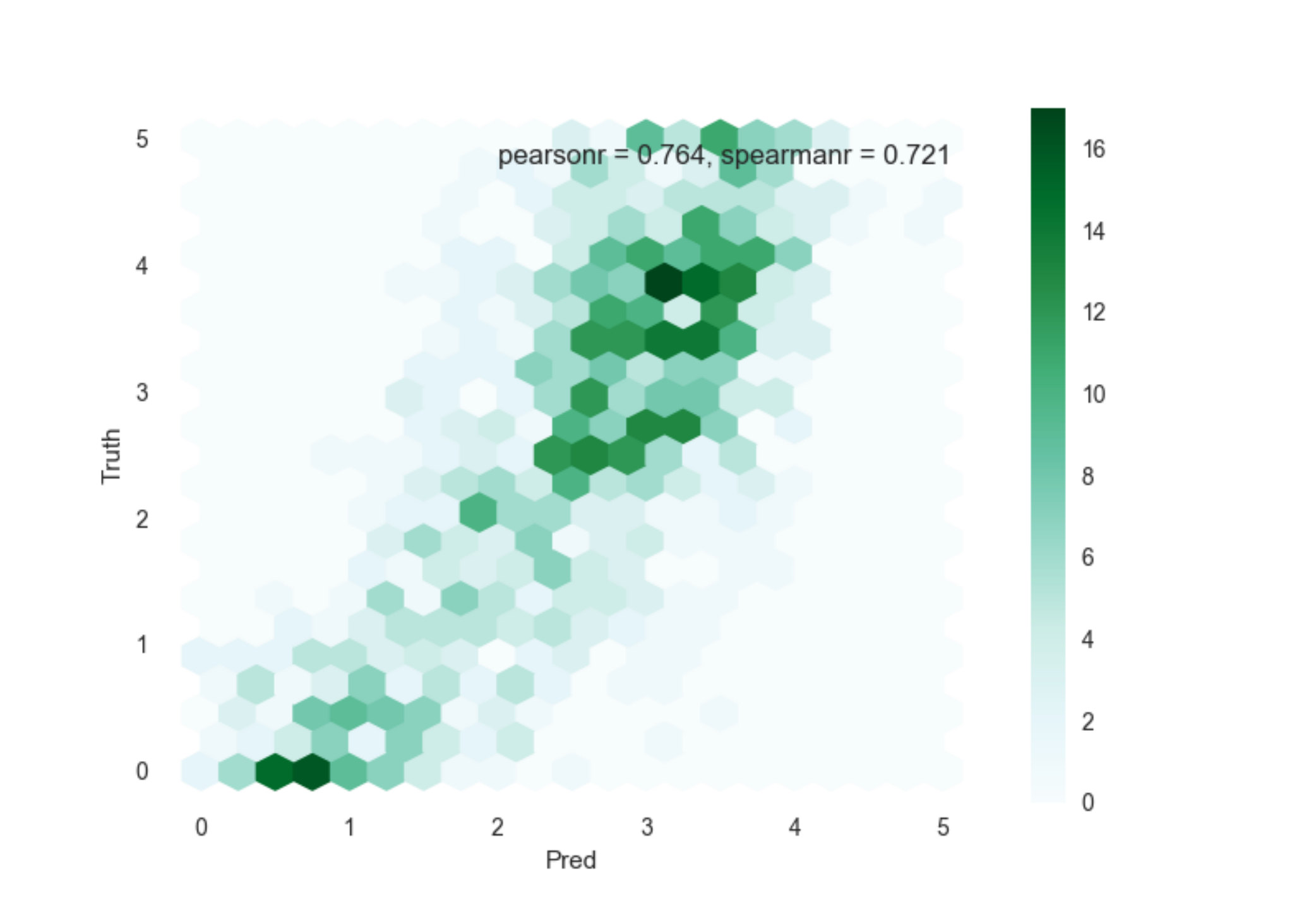}
		\caption{Two-layer BiRNN}
		\label{fig:STS_twobi}
	\end{subfigure}%
	\caption{Scatter plot for the STS Benchmark task result}\label{fig:STS_plot}
\end{figure}

We conducted the experiment in the same manner as for InferSent. For two sentence vectors $u$ and $v$, the component-wise product $u \cdot v$ and the absolute difference $\lvert u-v\rvert$ are computed and concatenated to be used as an input. As the target, the human rated similarity score $y$ is transformed as follows. Let $r^T=[1,...,5]$ denote a vector that takes integer values between 1 and 5. The target $y$ is transformed to the distribution $d$ using the equation below:
\begin{equation}
d_i=
\begin{cases}
y-\lfloor y \rfloor, & \text{if } i=\lfloor y \rfloor + 1,\\
\lfloor y \rfloor + 1, & \text{if } i=\lfloor y \rfloor,\\
0 & \text{otherwise}.
\end{cases}
\end{equation}
Finally, we trained a logistic regression model that predicts the transformed target $d$ from the sentence pair representations of the training dataset. The results for the STS Benchmark test dataset are summarized in Table \ref{table6.STS result}. Figure \ref{fig:STS_plot} presents a scatter plot of the results for the proposed models and the target $y$.

\begin{table}[t!]
	\begin{center}
		\caption{ STS Benchmark task dataset description} \label{table5.STS data description}%
		\renewcommand{\arraystretch}{1}
		\footnotesize{
		\centering{\setlength\tabcolsep{10pt}
			\begin{tabular}{c|c|c|c|c}
				\hline
				- & Train & Dev & Test & Total \\ 
				\hline
				\# of data &  5,749 &  1,500 &  1,379 &  8,628 \\ 
				\hline
		\end{tabular}}}
	\end{center}
\end{table}
\begin{table}[t!]
	\begin{center}
		\caption{Experimental results for the STS Benchmark task} \label{table6.STS result}%
		\renewcommand{\arraystretch}{1}
		\footnotesize{
			\centering{\setlength\tabcolsep{10pt}
				\begin{tabular}{c|c}
					\hline
					\multirow{2}{*}{\textbf{Model}} & \textbf{Pearson} \\
					& \textbf{correlation} \\ 
					\hline
					PV-DBOW  &  \multirow{3}{*}{0.649 } \\
					\cite{Doc2vec}  & \\
					\cite{lau2016empirical} & \\ 
					\hline
					SipThought & \multirow{2}{*}{0.721} \\
					\cite{SkipThought} & \\ 
					\hline
					SIF  & \multirow{2}{*}{0.720} \\
					\cite{SIF} & \\ 
					\hline
					Sent2vec &  \multirow{2}{*}{0.755} \\
					\cite{Sent2vec} & \\ 
					\hline
					InferSent & \multirow{2}{*}{0.758} \\
					\cite{InferSent} & \\ 
					\hline
					P-thought & \multirow{2}{*}{\textbf{0.812}} \\
					(one layer-Bi RNN) & \\ 
					\hline
					P-thought & \multirow{2}{*}{\textbf{0.797}} \\
					(two layers-Forward RNN) & \\ 
					\hline
					P-thought & \multirow{2}{*}{\textbf{0.764}} \\
					(two layers-Bi RNN) & \\ 
					\hline
		\end{tabular}}}
	\end{center}
\end{table}

The experimental results show that the P-thought models of all three levels outperformed the benchmarked models. An interesting observation is that the performances of the P-thought models for the STS benchmark task are inversely proportional to the model complexity: the simplest model (one-layer Bi RNN) yielded the highest correlation value, while the most complex model (two-layers Bi RNN) resulted in the lowest correlation value among the three P-thought models. This observation is exactly the opposite of the result for the MS-COCO dataset, where the more complex the P-thought model, the higher is the P-coherence score. One possible reason for this reversed performances is that the MS-COCO caption dataset used for the model training only contains around 600,000 sentences, which is far fewer than training datasets for general sequence learning tasks in the NLP field. Hence, it is more likely to overfit the training dataset for a more complex structure. This problem can be alleviated by obtaining more of paraphrase sentence pairs.

\section{Conclusion}

Sentence embedding is one of the most important text processing techniques in NLP. To date, various sentence embedding models have been proposed and have yielded good performances in document classification and sentiment analysis tasks. However, the fundamental ability of sentence embedding methods, i.e., how effectively the meanings of the original sentences are preserved in the embedded vectors, cannot be fully evaluated through such indirect methods.

In this study, under the proposition that a good sentence embedding method should act similar to human language recognition, we suggested the concept of semantic coherence and proposed a model named P-thought that aims to maximize the semantic coherence by designing a model to have a dual generation structure. The proposed model was evaluated based on the MS-COCO caption and STS Benchmark datasets. Experimental results showed that the P-thought models yielded better performances than the benchmarked models for both tasks. Based on the scatter plots in the two-dimensional space reduced by $t$-SNE, it can clearly be observed that the paraphrase sentences are more concentrated for the P-thought models than those using other sentence embedding methods. 
 	
The main limitation of the current work is that there are insufficient paraphrase sentences for training the models. P-thought models with more complex encoder structures tend to overfit the MS-COCO datasets. Although this problem can be resolved by acquiring more paraphrase sentences, it is not easy in practice to obtain a large number of paraphrase sentences. Therefore, similar to the approaches that have achieved good performances in machine translation by employing semi-supervised learning or unsupervised learning \cite{cheng2016semi,artetxe2017unsupervised,lample2017unsupervised}, an approach to improve the performances of the proposed models using only minimal paraphrase data should be developed.

\bibliographystyle{./sty/icml2017} 
\bibliography{References}
\end{CJK}
\end{document}